\documentclass[10pt,twocolumn]{article}
\usepackage{amsmath,graphicx}
\usepackage{spconf}
\usepackage{times}
\usepackage{epsfig}
\usepackage{graphicx}
\usepackage{color}
\usepackage{amsmath, dsfont, bbm, bm, amsfonts}
\usepackage[normalem]{ulem}
\usepackage{soul}
\usepackage{amssymb}
\usepackage[T1]{fontenc}
\usepackage{fixltx2e}
\usepackage{color}
\usepackage{algorithm}
\usepackage{algpseudocode}
\usepackage[font = small, justification=centering]{caption}
\usepackage{subcaption}
\usepackage{subcaption,placeins,blindtext}
\usepackage{setspace}
\usepackage[numbers]{natbib}
\usepackage{notoccite}

\usepackage{caption}

\def\bTheta{\boldsymbol\Theta}
\def\balpha{\boldsymbol\alpha}

\def\bx{\mathbf{x}}

\def\bX{\mathbf{X}}
\def\bd{\mathbf{d}}
\def\bD{\mathbf{D}}

\def\bB{\mathbf{B}}
\def\bW{\mathbf{W}}
\def\by{\mathbf{y}}

\def\bc{\mathbf{c}}
\def\bC{\mathbf{C}}

\def\bbR{{\mathbb R}}

\makeatletter
\newcommand*{\rom}[1]{\expandafter\@slowromancap\romannumeral #1@}
\makeatother

 \newcommand{\norm}[1]{\lVert #1 \rVert} 
\renewcommand\abstractname{\textit{Abstract}}

\makeatother

\usepackage[breaklinks=true,bookmarks=false]{hyperref}

\begin{document}

\title{Sparse Coding with Fast Image Alignment via Large Displacement Optical Flow}
%
\name{Xiaoxia Sun$^\dagger$, Nasser M. Nasrabadi$^\ddagger$ and Trac D. Tran$^\dagger$ \thanks{This work has been partially supported by NSF under Grants NSF-CCF-1117545, NSF-CCF-1422995 and NSF-EECS-1443936.}   }
\address{$^\dagger$ Department of ECE, Johns Hopkins University, 3400 N. Charles Street, Baltimore, MD, 21218 \\
$^\ddagger$ Lane Department of CSEE, West Virginia University,  395 Evansdale Drive, Morgantown, WV, 26506
}

\maketitle

\begin{abstract}
Sparse representation-based classifiers have  shown   outstanding accuracy and  robustness in image classification tasks even with the presence of intense noise and occlusion.  However, it has been discovered that the performance degrades significantly either when test image is not aligned with the dictionary atoms or the dictionary atoms themselves are not aligned with each other, in which cases  the sparse linear representation assumption fails.  In this paper, having both training and test images misaligned, we introduce  a novel sparse coding framework that is able to efficiently adapt the dictionary atoms to the test image  via large displacement optical flow.  In the proposed algorithm, every dictionary atom is automatically aligned with the input image and the sparse code is then recovered using the adapted dictionary atoms.    A corresponding supervised dictionary learning algorithm  is also developed for the proposed framework.  Experimental results on digit datasets recognition verify the efficacy and robustness of the proposed algorithm.
\end{abstract}




%

%
%

\section{Introduction}
Sparse coding has been successfully applied to numerous computer vision tasks, including face recognition \cite{jwright}, scene categorization  \cite{local_sparse_coding} and object detection \cite{Agarwal02}. Application of sparse representation-based classifier (SRC) on face recognition \cite{jwright} demonstrates a startling robustness over  noise and occlusions, where the test subjects are still recognizable even when they wear sunglasses or scarf.  However, SRC has been found to be highly sensitive to the misalignment of the image dataset: a small amount of image distortion due to translation, rotation, scaling and $3$-dimensional pose variations can lead to a significant degradation on the classification performance \cite{robustface}.

One straightforward way to solve the misalignment problem is to register the test image with dictionary atoms before sparse recovery. By assuming the dictionary atoms are registered, Wagner \emph{et al.} \cite{robustface} parameterize the misalignment of the test image with an affine transformation. These parameters are optimized using generalized Gauss-Newton methods after linearizing the affine transformation constraints. By minimizing the sparse registration error iteratively and sequentially for each class, their framework is able to deal with a large range of variations in translation, scaling, rotation and even $3$D pose variations. Due to the adoption of holistic features, sparse coding is more robust and less likely to overfit.

In the case of local feature-based sparse coding,  max pooling strategy \cite{honglaklee} is often employed over the neighboring coefficients to produce local translation-invariant property.    Based on spatial pyramid matching framework, Yang \emph{et. al.} \cite{local_sparse_coding} proposed a local sparse coding model with local SIFT features followed by multi-scale max pooling. The results on several large variance datasets achieved plausible performance that can hardly be pursued by simply applying holistic sparse coding. To improve the discriminability of the sparse codes, their dictionary was trained with supervised learning via backpropagation \cite{Yang}. Classification performance of local feature-based sparse coding has also been evaluated on several large datasets in \cite{ICML2011Coates_485}, demonstrating a state-of-art performance that is competitive with deep learning \cite{Hinton06afast}.  Another interesting approach is the convolutional sparse coding \cite{lecunn_conv_sparse_coding}, where the local features are reconstructed by convoluting the local sparse codes using local dictionary. Visualization of its dictionary shows that the dictionary atoms contain more complex features, therefore having more discriminative power. 
 
  \begin{figure*}[t]
\begin{center}
 \centerline{\includegraphics[width=1.0\linewidth]{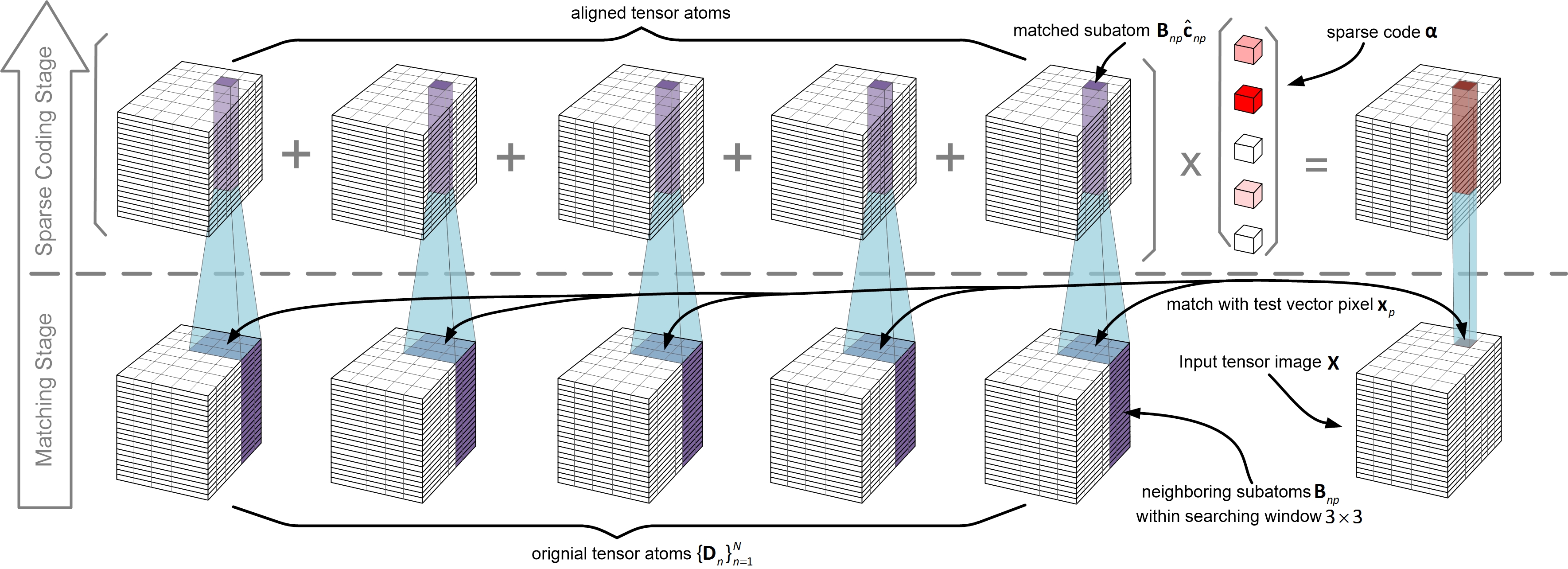}}
\end{center}
\captionsetup{justification=raggedright,
singlelinecheck=false
}
   \caption{Proposed sparse coding framework: Dictionary tensor atoms $\{\bD_n\}_{n=1}^{N}$ and the test tensor image $\bX$ are shown in the lower part of the figure. Searching window of size $T = 3\times 3$ within each tensor atom is colored with purple. Each group of neighboring $T$ subatoms $\bB_{np}$ is matched with the corresponding vector pixel $\bx_{p}$  of the test tensor image, resulting in an aligned subatom $\bB_{np}\hat\bc_{np}$.  After the matching process, the sparse code for $\bx_p$ is recovered using all the aligned subatoms $\{\bB_{np}\hat\bc_{np}\}_{n=1}^N$. For illustration purposes, only five  dictionary tensor atoms are shown in the figure and the magnitude of the sparse codes  are displayed with various intensities in red.}
\label{fig:pipeline}
\end{figure*}

In this paper, we present a novel sparse coding framework that is robust to image transformation. In the proposed model, each dictionary atom is constructed in the form of a tensor and is aligned with the test image using the large displacement optical flow concept \cite{large_optical_flow}.  We show experimentally that the proposed sparse coding framework outperforms most other sparsity-based methods.
 Specifically, our paper has the following novelties and contributions: {\it(\romannumeral1)} The proposed algorithm does not require the training dataset to be pre-aligned.
{\it(\romannumeral2)} Adapting the dictionary to the input test image is highly efficient: requiring only ${\cal{O}} (PT)$ operations for adapting each dictionary atom, where $T$ is the number of pixels in a searching window and $P$ is the total number of \emph{subatoms} to be aligned.
{\it(\romannumeral3)} Supervised dictionary learning algorithm is developed for the proposed sparse coding framework.

The remainder of the paper is organized as follows: We first introduce the proposed sparse coding framework for dealing with dataset misalignment in Section \ref{section:sc_imagealignment}. Next, in Section \ref{section:dictionary_learning}, we show how to train the dictionary in a supervised manner by solving a bilevel optimization problem. Finally, in Section \ref{section:experiments}, experimental results demonstrate that the proposed framework has a state-of-art performance, which is more promising over most existing sparsity-based methods.


\section{Sparse Coding with Image Alignment via Large Displacement Optical Flow}
\label{section:sc_imagealignment}

In this section, we first introduce how to construct the dictionary atoms and input images in the form of tensors.  We then illustrate how to eliminate the misalignment by dynamically adapt the tensor dictionary atoms to the input tensor image.

In the proposed sparse coding model, as shown in Fig. \ref{fig:pipeline}, both dictionary atom and input image are represented by image tensors. Each pixel in the tensor image is a vectorized version of a local patch in the original image, referred to as a vector pixel.  Denote the $n^{\text{th}}$ tensor atom as $\bD_n = [\bd_{n1},\dots, \bd_{nP}]\in\bbR^{M\times P}$ and a given test tensor image   as $\bX = [\bx_{1},\dots, \bx_P]\in \bbR^{M\times P}$, where $\bd_{np}\in\bbR^{M}$ is  the $p^{\text{th}}$ subatom of the $n^{\text{th}}$ tensor atom    and $\bx_{p}\in\bbR^M$ is the $p^{\text{th}}$ vector pixel of the input image. $M$ is the dimension of vector pixel, $n$ is the dictionary atom index and $P$ is the total number of subatoms in the tensor atom, which is the same number of vector pixels in the test tensor image. The dictionary is denoted as $\bD = [\bD_1, \dots, \bD_N] \in \bbR^{M\times NP}$.  Given  a dictionary with $N$ tensor atoms, a typical sparse recovery problem \cite{jwright} is formulated as:
\begin{align}
\hat\balpha = \arg\min_{\balpha } \frac{1}{2}\sum_{p=1}^P \norm{\sum_{n=1}^N \alpha_n \bd_{np}  - \bx_p}_2^2 + \lambda \norm{\balpha}_1,
\label{eq:model_sc}
\end{align}
where $\balpha = [\alpha_1, \dots, \alpha_N]^\top\in\bbR^N$ is the sparse coefficient and $\lambda>0$ is the regularization parameter.  
Problem (\ref{eq:model_sc}) is a standard form of $\ell_1$-sparse recovery problem that can be efficiently solved using alternating direction method of multipliers (ADMM) \cite{boyd}. 


%
When images in both the training and test datasets are misaligned, sparse coefficients recovered by solving the problem (\ref{eq:model_sc}) become unreliable, thus resulting in poor classification performance. To alleviate the misalignment problem, we propose to register each tensor atom with the input test image via large displacement optical flow \cite{large_optical_flow}. The notion of optical flow field is used here to describe the displacements of vector pixels within each tensor atom, and the sparse recovery is then performed by using only the best matching subatoms selected from the tensor atoms. The proposed framework is illustrated in Fig. \ref{fig:pipeline}.  Denote $\bB_{np}  \in\bbR^{M\times T}$ as the $T$ subatoms within the searching window centered at the location $p$ of the $n^{\text{th}}$ tensor atom. The recovery of the optical flow and sparse codes can be formally described as follows:
\begin{equation}
\begin{aligned}
&\left(\hat\balpha, \{\hat\bc_{np}\}\right) = \arg\min_{\balpha, \{\bc_{np}\} } \frac{1}{2}\sum_{p=1}^P \norm{ \sum_{n=1}^N \alpha_n\bB_{np} \bc_{np} - \bx_p}_2^2 + \lambda \norm{\balpha}_1, \\
&\text{s.t.} \quad \norm{\bc_{np}}_0 = 1, \norm{\bc_{np}}_1 = 1, \bc_{np}\geq \mathbf{0}, \\
& \quad\quad\; \forall n\in[N],p\in[P],
\label{eq:model_of1}
\end{aligned} 
\end{equation}
where $\norm{\bc_{np}}_0=1$ is the cardinality constraint and  $\bc_{np}\in \bbR^{T}$ is the sparse index vector that is used to characterize the optical flow field. The constraint in (\ref{eq:model_of1}) suggests that $\bc_{np}$  is a binary index vector and only one element is nonzero, which means that it can only select one subatom within the searching window.

The optimization problem in (\ref{eq:model_of1}) is a mixed-integer problem and NP-hard \cite{optimization_integers}.  Therefore, we propose a heuristic algorithm to find an informative $\balpha$ and the sparse index vectors $\{\bc_{np}\}_{n,p=1}^{N,P}$ for all vector pixels. As shown in Fig. (\ref{fig:pipeline}), the optical flow field for each vector pixel is found by searching for the best match between neighboring subatoms and the corresponding input vector pixel.  In practice, we found that searching for the best match without involving the sparse code is the key to render plausible performance in both classification accuracy and computational efficiency. Formally, we propose to find a local optimum of problem (\ref{eq:model_of1}) by solving the following optimization problem:
\begin{equation}
\begin{aligned}
&\hat\balpha = \arg\min_{\balpha} \frac{1}{2}\sum_{p=1}^P \norm{ \sum_{n=1}^N \alpha_n\bB_{np} \hat\bc_{np} - \bx_p}_2^2 + \lambda \norm{\balpha}_1  \\
&\text{s.t.} \quad \hat\bc_{np} = \arg\min_{\bc_{np}} \frac{1}{2}\norm{\bB_{np} \bc_{np} - \bx_p}_2^2, \\
&\quad\quad \norm{ \bc_{np} }_0= 1,  \norm{\bc_{np}}_1 = 1, \bc_{np}\geq \mathbf{0}, \\
&\quad\quad\; \forall n\in[N],p\in[P].
\label{eq:model_of2}
\end{aligned}
\end{equation}
In our approach, the sparse coding part of  (\ref{eq:model_of2}) is solved by using the alternating direction method of multipliers (ADMM) \cite{boyd}.  One important advantage of the above model is that it is highly  computational efficient because it only takes ${\cal{O}} (T)$ operations to search for the best match for each vector pixel. 


\section{Supervised Dictionary Learning}
\label{section:dictionary_learning}
In order to improve the efficiency of sparse coding and discriminablity of the dictionary, we employ the supervised dictionary learning framework \cite{Yang,bilevelsc,MairalTDDL} to optimize the dictionary and the classifier parameters simultaneously. Formulated as a bilevel optimization problem, the dictionary is updated using back propagation to minimize the classification error. Formally, the supervised dictionary learning problem can be formulated as follows:
\begin{align}
\min\limits_{\bW, \bD}\mathbb{E}_{\by,\bX}\left[\ell(\by, \bW\hat\balpha(\bX, \{\hat\bc_{np}(\bD)\}, \bD))\right] + \frac{\mu}{2}\norm{\bW}_F^2,
\label{eq:dl2}
\end{align}
where $\ell(\cdot)$ is some smooth and convex function that is used to define the classification error and $\mu>0$ is the regularization parameter used to alleviate the overfitting of the classifier. Due to the triviality of updating classifier parameters, here we only state the update for the dictionary:
\begin{align}
\bD\leftarrow \Pi(\bD - \rho^t\cdot \partial \ell / \partial \bD),
\end{align}
where $\rho>0$ is the learning rate, $t$ is the iteration counter and $\Pi$ is the projection that regulate the Frobenius norm of every tensor atom  to be one.  Similar to \cite{Yang,bilevelsc,MairalTDDL}, (\ref{eq:dl2}) suggests that the update of both the dictionary and the classifier are driven by reducing classification error. The local optima can be solved by using descent method \cite{bilevel_survey} based on error backpropagation.   
The sparse code $\balpha$ is an implicit function of $\bX, \{\bc_{np}\}$ and $\bD$. In addition, each optical flow field $\bc_{np}$ is an implicit function of $\bD$ and $\bx_{np}$. Therefore, given an input image $\bX$ and an optimal sparse code $\hat\balpha$, apply the chain rule of differentiation, the direction along which the upper-level cost decreases can be formulated as:
\begin{align}
\frac{\partial \ell(\by, \bW\balpha)}{\partial \bD} = \frac{\partial \ell}{\partial \balpha}\frac{\partial \balpha}{ \partial \bD} + \sum_{p=1}^P\frac{\partial \ell}{\partial \bC_p}  \frac{\partial \bC_{p}}{\partial \bD},
\label{eq:chainrule}
\end{align}
where $\bC_p =  \bigoplus_{n=1}^N \bar\bc_{np} \in \bbR^{NP\times N}$ and $\bigoplus$ denotes the direct sum. Also, $\bar\bc_{np}\in\bbR^{NP}$ is obtained by zero-padding with $\bc_{np}$, where $(N-1)P+1$ to $NP$ elements of $\bar\bc_{np}$ are from those of $\bc_{np}$. Due to the binary constraints on $\{\bc_{np}\}$, every element of the gradient $\partial \bC_{p}/\partial \bD$ equals to zero. On the other hand, the first part of the derivative can be solved by applying fixed point differentiation \cite{differentiablesparse}. Due to the page limitation  of the paper and the triviality for deriving the term $\partial\ell/\partial\balpha$, we only show the final derivation of $\partial \balpha/\partial \bD$ as follows:
\begin{equation}
\frac{\partial \balpha_\Lambda}{\partial d_{mnp}} = \bTheta_{\Lambda,\Lambda}^{-1} \left( \frac{\partial {(\bD\bC_p)}_\Lambda^\top}{\partial d_{mnp}}\bx_p -\frac{\partial \bTheta_{\Lambda,\Lambda}}{\partial d_{mnp}}\balpha_\Lambda \right),
\label{eq:dl_alpha_d}
\end{equation}
 where $\Lambda$ is the index set of active atoms of the sparse code $\balpha$. ${(\bD\bC_p)}_\Lambda$ is the matrix obtained by collecting the active columns of ${\bD\bC_p}$,  $\bTheta = \sum_{\rho=1}^P \bC^{\top}_p\bD^{\top}\bD\bC_p$ and $\bTheta_{\Lambda,\Lambda}$ is the submatrix obtained by selecting the active columns and rows of $\bTheta$. The matrix $\bTheta_{\Lambda,\Lambda}$ is always nonsingular since the total number of measurement $MP$ is always significantly larger than the number of active atoms. Combining  (\ref{eq:chainrule}) with  (\ref{eq:dl_alpha_d}) for each dictionary element, the gradient for updating the dictionary can be achieved. For a large dataset, the dictionary and the classifier parameters are updated in an online manner. 

	  \begin{figure*}[ht]
	  \centering
	  \begin{subfigure}[b]{.2\textwidth}
	        \centering
	         \centerline{       \includegraphics[width=4.5cm]{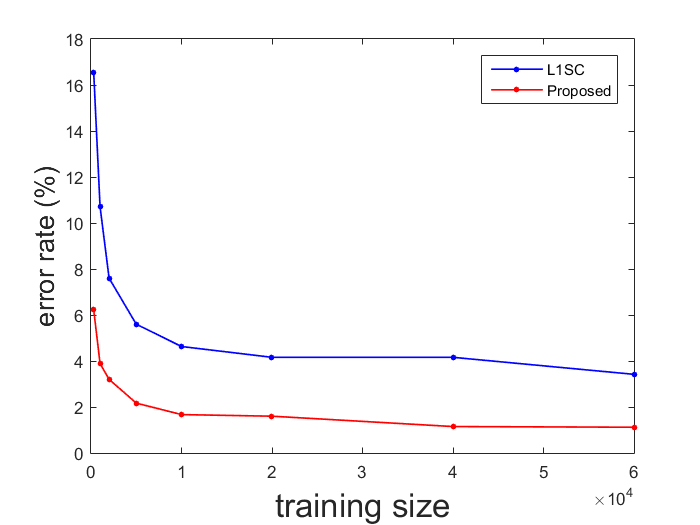}  }
	                \caption{}
	                 \label{fig:subsample}
	  \end{subfigure} \hspace{2em}
	  \begin{subfigure}[b]{.2\textwidth}
	        \centering
	         \centerline{       \includegraphics[width=4.5cm]{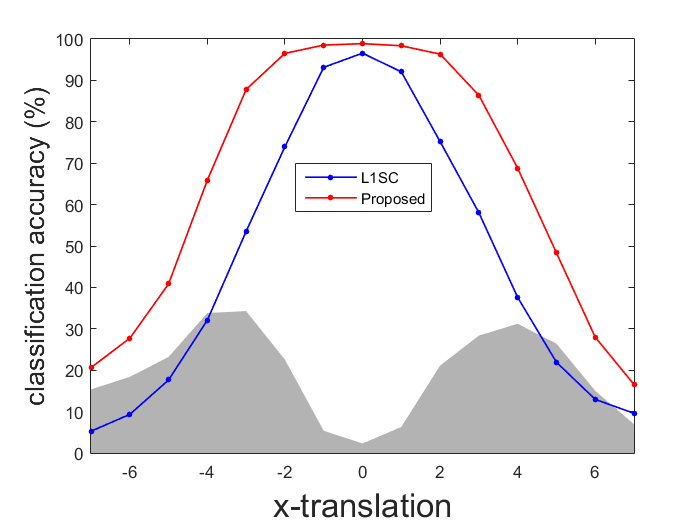}  }
	                \caption{}
	                 \label{fig:translation}
	  \end{subfigure}  \hspace{2em}
	  \begin{subfigure}[b]{0.2\textwidth}
	        \centering
	      \centerline{         \includegraphics[width=4.5cm]{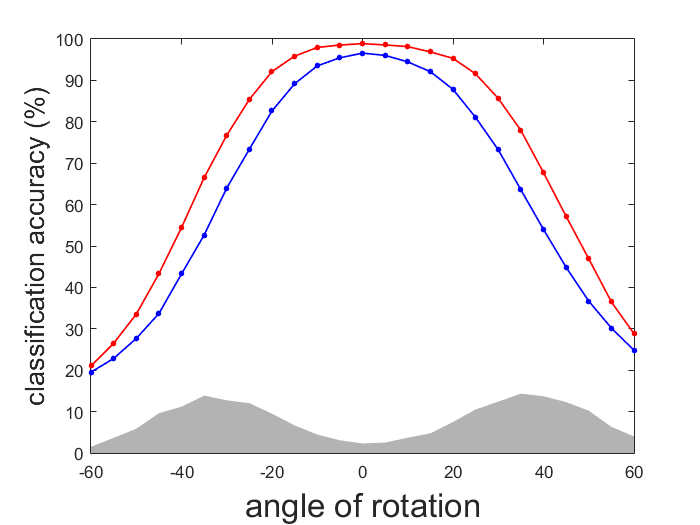}
	      }
	                \caption{}
	                \label{fig:scaling}
	  \end{subfigure}  \hspace{2em}
	  \begin{subfigure}[b]{0.2\textwidth}
	        \centering
	           \centerline{     \includegraphics[width=4.5cm]{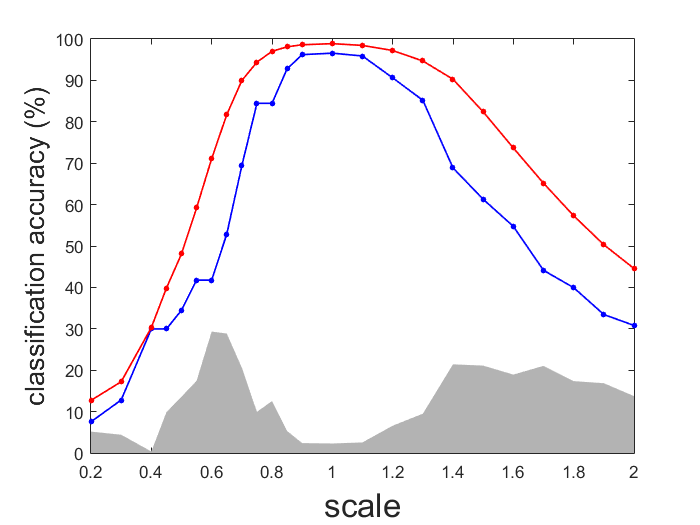}
	           }
	                \caption{}
	                \label{fig:rotation}
	  \end{subfigure}
	\captionsetup{justification=raggedright, singlelinecheck=false}
	
	\caption{\footnotesize The proposed method demonstrates plausible performance on MNIST digits recognition  with a small number  of training samples. It also demonstrates robustness towards  various  image deformations. Classification accuracy of different experimental settings are shown in the above sub-figures: (a) Error rate under various sizes of training samples. (b)  Translation along $x$ direction versus classification accuracy. (c) In-plane rotation only. (d) Scale variation only. In (b)-(c), red and blue lines are the results of the proposed method and L1SC, respectively. Gray shadow area at the bottom of each figure is the accuracy difference between the proposed method and L1SC. }
	\label{fig:invariance}
	\end{figure*}

\section{Experiments}
\label{section:experiments}
In this section, we evaluate the proposed algorithm on hand-written digits datasets including the MNIST and USPS.   The sparse coding is performed with a single dictionary and  linear SVM is used for classification. For a fair comparison, we only compare with the results that are produced with the same SRC strategy. The dictionary size in our paper is set to be no larger than those used in other methods. Similar to \cite{Yang}, parameters in our experiments are chosen heuristically. The batch size for updating the dictionary is $512$. Initial learning rate $\rho$ is set to $0.001$ and $\lambda=0.01$.

\subsection{Evaluation on the MNIST Database}
MNIST \cite{MNIST}  consists of a total number of $70,000$ images of digits, of which $60,000$ are training set and the rest $10,000$ are test set. Each digit is centered and normalized in a $28\times28$ field.  The dictionary size $N$  is set to be $150$ for this database. 

We first evaluate the performance of the proposed algorithm under various number of training samples. We follow the same experimental setting as in \cite{ckn}, examining the classification accuracy given the training size  $\{300, 1K, 2K, 5K, 10K, 20K, 40K, 60K\}$. The performance is shown in Fig. \ref{fig:invariance} (a).  The proposed method significantly outperforms the $\ell_1$ sparse coding-based algorithm (L1SC) \cite{MairalTDDL}. 

We then demonstrate the robustness of the proposed method towards various image deformations. Following a similar setting as in \cite{robustface},  we perform the translation along $x$ direction, rotation and scaling separately only on the test samples. We report the classification accuracy with respect to various levels of deformation and compare the performance with L1SC. The experimental results are shown in Fig. \ref{fig:invariance}(b)-(d). Performance of our method and L1SC are illustrated in red and blue lines, respectively. The shadow area at the bottom of each figure is the accuracy difference  between the two methods. We can see for all three deformations, the proposed method consistently outperforms L1SC. In addition, the hump shape of the shadow area indicates that the proposed method is robust to numerous image deformations.

Finally, the error rate for the MNIST is shown in Table \ref{table:oa}. Our method reaches the lowest error rate of $1.12\%$. On MNIST, differences of more than $0.1\%$ are statistically significant \cite{greedy_dnn}. Comparing with the second best algorithm, the proposed method reduces the error rate by $0.12\%$, exhibiting better generality and dictionary compactness.

\subsection{Evaluation on the USPS Database}
The USPS dataset has $7,291$ training and $2,007$ test images, where each of them is of size $16\times 16$. Being compared to MNIST, the USPS dataset has a much larger variance and a smaller training set, which challenges the dictionary generality. 
For a fair comparison, the dictionary size $N$ is set to be $80$. Local patch size is $5\times 5$ ($M$ = $25$). Searching window size is $5\times 5$  ($T$ = $25$). The performance of various approaches on USPS database are depicted in Table \ref{table:oa}.    Our algorithm achieves the lowest error rate $3.43\%$ among other supervised learning-based methods. The experimental result validates the efficacy of our proposed algorithm on a dataset with a larger variance.

	\begin{table}
\begin{center}
\begin{tabular}{|l|c|c|}
\hline
Method & MNIST  & USPS  \\
\hline\hline
CBN &$1.95$ $(3\times10^4)$ &$4.14$ $(7291)$ \\
ESC \cite{ffs} &$5.16$ $(150)$  &$6.03$ $(80)$\\
Ramirez \emph{et al.} \cite{incoh_dl}  & $1.26$ $(800)$  &$3.98$ $(80)$\\
Deep Belief Network \cite{Hinton06afast} &$1.25$ $($-$)$ & - $($-$)$\\
MMDL \cite{zhaowen} &$1.24$ $(150)$ &-$($-$)$\\
Proposed &$\mathbf{1.12}$ $(150)$ &$\mathbf{3.43}$ $(80)$\\
\hline
Improvements &$9.7\%$ &$13.8\%$ \\
\hline
\end{tabular}
\end{center}
	\captionsetup{justification=raggedright, singlelinecheck=false}
\caption{ Error Rate (\%) on MNIST and USPS datasets. The dictionary size is shown in the parentheses. Improvements over the second best algorithm is shown in the last line.}
\label{table:oa}
\end{table}


\section{Conclusion}
\label{section:conclusion}
In this paper, we present a novel sparse coding algorithm that is able to dynamically select the dictionary subatoms to adapt to the misaligned image dataset. In the proposed method, both the dictionary atoms and the input test image are represented by tensors, and each vector pixel in the tensor image is a vectorized local patch. Each tensor atom is aligned with the input tensor image using large displacement optical flow, which is highly computationally efficient. Using the fixed point differentiation, a supervised dictionary learning algorithm is developed for the proposed sparse coding framework, which significantly reduces the required dictionary size. 

\newpage

{
\bibliographystyle{ieee}
\bibliography{refs}

\begin{thebibliography}{10}

\bibitem{jwright}
J.~Wright, A.Y. Yang, A.~Ganesh, S.S. Sastry, and Y.~Ma,
\newblock ``Robust face recognition via sparse representation,''
\newblock {\em IEEE Trans. on Pattern Anal. and Mach. Intell.}, vol. 31, no. 2,
  pp. 210--227, Feb. 2009.

\bibitem{local_sparse_coding}
J.~Yang, K.~Yu, Y.~Gong, and T.~Huang,
\newblock ``Linear spatial pyramid matching using sparse coding for image
  classification,''
\newblock {\em in CVPR}, pp. 1794--1801, Jun. 2009.

\bibitem{Agarwal02}
S.~Agarwal and D.~Roth,
\newblock ``Learning a sparse representation for object detection,''
\newblock {\em in ECCV}, vol. 4, pp. 113--130, May 2002.

\bibitem{robustface}
A.~Wagner, J.~Wright, A.~Ganesh, Z.~Zhou, H.~Mobahi, and Y.~Ma,
\newblock ``Toward a practical face recognition system: Robust alignment and
  illumination by sparse representation,''
\newblock {\em IEEE Trans. on Pattern Anal. and Mach. Intell.}, vol. 34, no. 2,
  pp. 372--386, Feb. 2012.

\bibitem{honglaklee}
H.~Lee, R.~B. Grosse, R.~Ranganath, and A.~Y. Ng,
\newblock ``Convolutional deep belief networks for scalable unsupervised
  learning of hierarchical representations,''
\newblock {\em in ICML}, Jun. 2009.

\bibitem{Yang}
J.~Yang, K.~Yu, and T.~Huang,
\newblock ``Supervised translation-invariant sparse coding,''
\newblock {\em in CVPR}, pp. 3517--3524, Jun. 2010.

\bibitem{ICML2011Coates_485}
A.~Coates and A.~Y. Ng,
\newblock ``The importance of encoding versus training with sparse coding and
  vector quantization,''
\newblock {\em in ICML}, pp. 921--928, Jul. 2011.

\bibitem{Hinton06afast}
G.~E. Hinton and S.~Osindero,
\newblock ``A fast learning algorithm for deep belief nets,''
\newblock {\em Neural Computation}, vol. 18, no. 7, pp. 527--1554, Jul. 2006.

\bibitem{lecunn_conv_sparse_coding}
K.~Kavukcuoglu, P.~Sermanet, Y.~Boureau, K.~Gregor, M.~Mathieu, and Y.~Lecun,
\newblock ``Learning convolutional feature hierarchies for visual
  recognition,''
\newblock {\em in NIPS}, pp. 1090--1098, Dec. 2010.

\bibitem{large_optical_flow}
T.~Brox and J.~Malik,
\newblock ``Large displacement optical flow: Descriptor matching in variational
  motion estimation,''
\newblock {\em IEEE Trans. on Pattern Anal. and Mach. Intell.}, vol. 33, no. 3,
  pp. 500--513, Mar. 2011.

\bibitem{boyd}
S.~Boyd, N.~Parikh, E.~Chu, B.~Peleato, and J.~Eckstein,
\newblock ``Distributed optimization and statistical learning via the
  alternating direction method of multipliers,''
\newblock {\em Journal FTML}, vol. 3, no. 1, pp. 1--122, Jan. 2011.

\bibitem{optimization_integers}
D.~Bertsimas and R.~Weismantel,
\newblock ``Optimization over integers,''
\newblock {\em Athena Scientific}, 2005.

\bibitem{bilevelsc}
J.~Yang, Z.~Wang, Z.~Lin, X.~Shu, and T.~Huang,
\newblock ``Bilevel sparse coding for coupled feature spaces,''
\newblock {\em in CVPR}, pp. 2360--2367, Jun. 2012.

\bibitem{MairalTDDL}
J.~Mairal, F.~Bach, and J.~Ponce,
\newblock ``Task-driven dictionary learning,''
\newblock {\em IEEE Trans. on Pattern Anal. and Mach. Intell.}, vol. 34, no. 4,
  pp. 791--804, Apr. 2012.

\bibitem{bilevel_survey}
B.~Colson, P.~Marcotte, and G.~Savard,
\newblock ``An overview of bilevel optimization,''
\newblock {\em Ann. of Operat. Res.}, vol. 153, no. 1, pp. 235--256, Apr. 2007.

\bibitem{differentiablesparse}
D.~M. Bradley and J.~A. Bagnell,
\newblock ``Differentiable sparse coding,''
\newblock {\em in NIPS}, Dec. 2008.

\bibitem{MNIST}
Y.~Lecun, L.~Bottou, Y.~Bengio, and P.~Haffner,
\newblock ``Gradient-based learning applied to document recognition,''
\newblock {\em Proceedings of the IEEE}, vol. 86, no. 11, pp. 2278--2324, Nov.
  1998.

\bibitem{ckn}
J.~Mairal, P.~Koniusz, Z.~Harchaoui, and C.~Schmid,
\newblock ``{Convolutional Kernel Networks},''
\newblock {\em in NIPS}, 2014.

\bibitem{greedy_dnn}
Y.~Bengio, P.~Lamblin, D.~Popovici, and H.~Larochelle,
\newblock ``Greedy layer-wise training of deep networks,''
\newblock {\em in NIPS}, pp. 153--160, Dec. 2007.

\bibitem{ffs}
H.~Lee, A.~Battle, R.~Raina, and A.~Y. Ng,
\newblock ``Efficient sparse coding algorithms,''
\newblock {\em in NIPS}, pp. 801--808, Dec. 2006.

\bibitem{incoh_dl}
I.~Ramirez, P.~Sprechmann, and G.~Sapiro,
\newblock ``Classification and clustering via dictionary learning with
  structured incoherence and shared features,''
\newblock {\em in CVPR}, pp. 3501--3508, Jun. 2010.

\bibitem{zhaowen}
Z.~Wang, J.~Yang, N.~M. Nasrabadi, and T.~Huang,
\newblock ``A max-margin perspective on sparse representation-based
  classification,''
\newblock {\em in ICCV}, pp. 1217--1224, Dec. 2013.

\end{thebibliography}
}

\end{document}